\documentclass[11pt,a4paper]{article}
\usepackage[hyperref]{naaclhlt2018}
\usepackage{times}
\usepackage{latexsym}
\usepackage{url}

\usepackage{amsmath}
\usepackage{multirow}

\aclfinalcopy 

\usepackage{graphicx}
\usepackage{subfig}
\usepackage{multicol}
\usepackage{diagbox}
\usepackage[textsize=small]{todonotes}
\usepackage{amstext}
\usepackage{covington}

\title{
    Behavior Analysis of NLI Models:  \\
    Uncovering the Influence of Three Factors on Robustness
}

\author{V. Ivan Sanchez Carmona \and Jeff Mitchell \and Sebastian Riedel \\
        University College London \\ Department of Computer Science \\ 
        \tt \{i.sanchezcarmona, j.mitchell, s.riedel\}@cs.ucl.ac.uk}

\date{}

\begin{document}
\maketitle
\begin{abstract}
  Natural Language Inference is a challenging task that has received substantial attention, and state-of-the-art models now achieve impressive test set performance in the form of accuracy scores. Here, we go beyond this single evaluation metric to examine robustness to semantically-valid alterations to the input data.
  We identify three factors - \emph{insensitivity}, \emph{polarity} and \emph{unseen pairs} - and compare their impact on three SNLI models under a variety of conditions.
  Our results demonstrate a number of strengths and weaknesses in the models' ability to generalise to new in-domain instances.
  In particular, while strong performance is possible on unseen hypernyms, unseen antonyms are more challenging for all the models. 
  More generally, the models suffer from an insensitivity to certain small but semantically significant alterations, and are also often influenced by simple statistical correlations between words and training labels. Overall, we show that evaluations of NLI models can benefit from studying the influence of factors intrinsic to the models or found in the dataset used.
  
\end{abstract}

\section{Introduction}
\label{section:introduction}

The task of Natural Language Inference (NLI)\footnote{Also known as Recognizing Textual Entailment.} has received a lot of attention and has elicited models which have achieved impressive results on the Stanford NLI (SNLI) dataset \cite{snli:emnlp2015}. Such results are impressive due to the linguistic knowledge required to solve the task \cite{LoBue:2011:TCK:2002736.2002805,NLI}. However, the ever-growing complexity of these models inhibits a full understanding of the phenomena that they capture. 

As a consequence, evaluating these models purely on test set performance may not yield enough insight into the complete repertoire of abilities learned and any possible abnormal behaviors \cite{kummerfeld-EtAl:2012:EMNLP-CoNLL,Sammons:2010:ATE:1858681.1858803}. A similar case can be observed in models from other domains; take as an example an image classifier that predicts based on the image's background rather than on the target object \cite{zhao-EtAl:2017:EMNLP20173,Ribeiro:2016:WIT:2939672.2939778}, or a classifier used in social contexts that predicts a label based on racial attributes \cite{crawford2016there}. In both examples, the models exploit a bias (an undesired pattern hidden in the dataset) to enhance accuracy. In such cases, the models may appear to be \emph{robust} to new and even challenging test instances; however, this behavior may be due to spurious factors, such as biases. Assessing to what extent the models are robust to these contingencies just by looking at test accuracy is, therefore, difficult.

In this work we aim to study how certain factors affect the robustness of three pre-trained NLI models (a conditional encoder, the DAM model \cite{parikh-EtAl:2016:EMNLP2016}, and the ESIM model \cite{chen-EtAl:2017:Long3}). We call these target factors \emph{insensitivity} (not recognizing a new instance), \emph{polarity} (a word-pair bias), and \emph{unseen pairs} (recognizing the semantic relation of new word pairs). We became aware of these factors based on an exploration of the models' behavior, and we hypothesize that these factors systematically influence the behavior of the models.

In order to systematically test if the above factors affect robustness, we propose a set of challenging instances for the models: We sample a set of instances from SNLI data, we apply a transformation on this set that yields a new set of instances, and we test both how well the models classify these new instances and whether the target factors influence the models' behavior. The transformation (swapping a pair of words between premise and hypothesis sentences) is intended to yield both easy and difficult instances to challenge the models, but easy for a human to annotate them.

We draw motivation to study the robustness of NLI models from previous work on evaluating complex models \cite{isabelle-cherry-foster:2017:EMNLP2017,white-EtAl:2017:I17-1}. Furthermore, we base our approach on the discipline of behavioral science which provides methodologies for analyzing how certain factors influence the behavior of subjects under study \cite{Epling1986}. 

We aim to answer the research questions: How robust is the predictive behavior of the pre-trained models under our transformation to input data? Do the target factors (insensitivity, polarity, and unseen pairs) influence the prediction of the models? Are these factors common across models? 

Our results show that the models are robust mainly where the semantics of the new instances do not change significantly with respect to the sampled instances and thus the class labels remain unaltered; i.e., the models are insensitive to our transformation to input data. However, when the class labels change, the models significantly drop accuracy. In addition, the models exploit a bias, polarity, to stay \emph{robust} when facing new instances. We also find that the models are able to cope with unseen word pairs under a hypernym relation, but not with those under an antonym relation, suggesting their inability to learn a symmetric relation.

\section{Related Work}

\subsection{Analysis of Complex Models}

Previous works in ML and NLP have analyzed different aspects of complex models using a variety of approaches; for example, understanding input-output relationships by approximating the local or global behavior of the model using an interpretable model \cite{Ribeiro:2016:WIT:2939672.2939778,craven1996extracting}, or analyzing the output of the model under lesions of its internal mechanism \cite{DBLP:journals/corr/LiMJ16a}. Another line of work has analyzed the robustness of NLP models both via controlled experiments to complement the information from the test set accuracy and test abilities of the models \cite{isabelle-cherry-foster:2017:EMNLP2017,bhashemi-hwa:2016:EMNLP2016,white-EtAl:2017:I17-1} and via adversarial instances to expose weaknesses \cite{jia-liang:2017:EMNLP2017}. In addition, work has been done to uncover and diminish \emph{gender} biases in datasets captured by structured prediction models \cite{zhao-EtAl:2017:EMNLP20173} and word embeddings \cite{bolukbasi-nips16}. However, to the best of our knowledge, there is no previous work to study the robustness of NLI models while analyzing factors affecting their predictions.

\subsection{Behavior Analysis}

Previous work on behavioral science has focused on understanding how environmental factors influence behaviors in both human \cite{soman2001effects} and animal \cite{doi:10.1093/ilar.39.1.20} subjects with the objective of predicting behavioral patterns or analyzing environmental conditions. This methodology also helps to identify and understand abnormal behaviour by collecting behavioral data without the need to reach any internal component of the subject \cite{birkett2011abnormal}.

We base our approach in the discipline of behavioral science since some of our research questions and objectives align to those from this discipline; in addition, its methodology to study how factors effect on the subjects' behavior provides statistical guarantees.

\section{Background}

\subsection{Natural Language Inference}
NLI, or RTE, is the task of inferring whether a natural language sentence (hypothesis) is entailed by another natural language sentence (premise) \cite{NLI,dagan_dolan_magnini_roth_2009,dagan_2004}. More formally, given a pair of natural language sentences $i=(premise, hypothesis)$, a model classifies the type of relation such sentences fall in from three possible classes,  \emph{entailment}, where the hypothesis is necessarily true given the premise, \emph{neutral}, where the hypothesis may be true given the premise, and \emph{contradiction}, where the hypothesis is necessarily false given the premise. Solving this task is challenging since it requires linguistic and semantic knowledge, such as co-reference, hypernymy, and antonymy \cite{LoBue:2011:TCK:2002736.2002805}, as well as pragmatic knowledge and informal reasoning \cite{NLI}. 

\subsection{Behavior Analysis}

Behavior analysis seeks to account for the role that factors (independent variables) play in the behavior (dependent variable) of subjects. Testing for the influence of a factor on the subject's behavior can be done via statistical tests: A null hypothesis states no association between a target factor and behavior, whereas the alternative hypothesis states an association \cite{mcdonald-2014}.

\section{Dataset and Models}

\subsection{SNLI Dataset}

The Stanford NLI dataset \cite{snli:emnlp2015} was created with the purpose of training deep neural models while providing human-annotated data. Each instance was created by providing a premise sentence, harvested from a pre-existing dataset, to a crowdsource worker who was instructed to produce three hypothesis sentences, one for each NLI class (entailment, neutral, contradiction). This process yielded a balanced dataset containing around 570K instances.

\subsection{Models}

\paragraph{Conditional Encoder} 
We use two bidirectional LSTMs; the first LSTM encodes the premise sentence into a fixed-size vector embedding by sequentially reading on a word basis, while the second LSTM encodes the hypothesis sentence conditioned on the representation of the premise sentence. At the final layer we used a softmax over the class labels on top of a 3-layer MLP. All embeddings, of dimensionality $d=100$, were randomly initialized and learned during training. Accuracy on SNLI's dev set is 0.782.

\paragraph{Decomposable Attention Model}

DAM \cite{parikh-EtAl:2016:EMNLP2016} consists of 2-layer multilayer-perceptrons (MLPs) factorized in a 3-step process. First, a soft-alignment matrix is created for all the words in both the premise and hypothesis. Then, each word of the premise is paired with the soft-alignment representation of the hypothesis sentence and fed into an MLP, and similarly for each word in the hypothesis with the soft-alignment of the premise. The resulting representations are then aggregated where the vector representations of the premise are summed up and the same for those of the hypothesis; the new representations are then fed to an MLP, followed by a linear layer and a softmax whose output is a class label. We use $d=300$ dimensional GloVe embeddings (not updated at training time). All layers use the ReLU function. Accuracy on SNLI's dev set is 0.854.

\paragraph {Enhanced Sequential Information Model}

ESIM \cite{chen-EtAl:2017:Long3} performs inference in three stages. First, Input Encoding uses BiLSTMs to produce representations of each word in its context within premise or hypothesis. Then, Local Inference Modelling constructs new word representations for each hypothesis (premise) by summing over the BiLSTM hidden states for the premise (hypothesis) words using weights from a soft attention matrix. Additionally, these representations are enhanced with element-wise products and differences of the original hidden states vectors and the new attention based vectors. Finally, Inference Composition uses a BiLSTM, average and max pooling and an MLP output layer to produce predicted labels. Accuracy on SNLI's dev set is 0.882.


\section{Methods}
\label{section:InterSentence_Word_Pair_Interaction}

We test our main hypothesis (Section~\ref{section:introduction}) by perturbing instances in a controlled, simple, and meaningful way. This alteration, at the instance level, yields new sets of instances which range from \emph{easy} (the semantics and the label of the new instance are the same to those of the original instance) to \emph{challenging} (both semantics and label of the new instance change with respect to those of the original instance), but all of them remain easy to annotate for a human.

To examine how the models generalize from seen instances to transformed instances, we sample our original instances from the SNLI training set, which we refer to as control instances from now on. We then produce new instances which differ either minimally from the control instances, by changing only a single word in the premise and hypothesis, or more substantially, by copying the same sentence structure into the premise and hypothesis with a single word changed.
In this way, we produce instances that contain only words seen at training time, 
within sentence structures also seen at training time. 
Thus, our evaluation sets are as in-domain as possible, 
and control for factors associated with novel sentential contexts and vocabulary.

\subsection{Basic Procedure and Statistical Analyses}

We first sample an instance from the SNLI dataset according to a given criterion, namely we look for a specific word pair in the instance; then, we apply our transformation over the word pair. This procedure generates a new instance. After that, the models label the new instance, and we statistically analyze which target factors influenced the models to respond in such a way via chi-square (McNemar's, independence, and homogeneity) tests  \cite{mcdonald-2014,Alpaydin:2010:IML:1734076}. When the sample size is too small we apply Yate's correction or a Fisher test. We use the StatsModels \cite{seabold2010statsmodels} and SciPy \cite{4160250} packages. The level of significance is $p<0.0001$, unless otherwise stated.\footnote{We apply a Bonferroni correction.} This procedure is applied in four experiments, where we study the effect of different word pairs (hypernym, hyponym, and antonyms) and the effect of two types of context words surrounding the word pairs which we refer to as \emph{in situ} and \emph{ex situ} (explained in Section \ref{section:experimental_conditions}).

\begin{table*}[ht]
\centering
\resizebox{\textwidth}{!}{%
\begin{tabular}{ c c  c  c c  c c c  c c c  c c c } 
 \hline
        & & \multicolumn{3}{c}{Whole sample}  & \multicolumn{3}{c}{Subset 1:} & \multicolumn{3}{c}{Subset 2:} & \multicolumn{3}{c}{Subset 3:} \\[0.5ex]
        &  &  &  &  & \multicolumn{3}{c}{Gold label changes} & \multicolumn{3}{c}{Unseen word pairs} & \multicolumn{3}{c}{Polarity $\neq$ gold label} \\[0.5ex]
        Exp & sample  & ESIM & DAM & CE & ESIM & DAM & CE & ESIM & DAM & CE & ESIM & DAM & CE \\
 \hline\hline
 1 & $\mathit{I_A}$ & 0.970 & 0.946 & 0.820 &  &  &  &  &  &  & 0.900 & 0.900 & 0.750 \\ 
 & $\mathit{I_{TA1}}$ & 0.933 & 0.946 & 0.732 &  &  &  & 0.600 & 0.500 & 0.400 & 0.681 & 0.637 & 0.536 \\
 & $\mathit{I_{TA2}}$ & 0.721 & 0.771 & 0.645 & 0.554 & 0.653 & 0.476 &   &  &  &  &  &  \\
 & $\mathit{I_{TA3}}$ & 0.722 & 0.745 & 0.646 & 0.568 & 0.630 & 0.535 &  &  &  &  &  &  \\
 2 & $\mathit{E_A}$ & 0.953 & 0.958 & 0.508 &  &  &  &  &  &  & 0.400 & 0.500 & 0.450 \\
 & $\mathit{E_{TA}}$ & 0.933 & 0.929 & 0.480 &  &  &  & 0.575 & 0.500 & 0.175 & 0.565 & 0.492 & 0.260 \\
 3 & $\mathit{I_H}$ & 0.898 & 0.819 & 0.828 &  &  &  &  &  &  & 0.836 & 0.701 & 0.733 \\ 
 & $\mathit{I_{TH}}$ & 0.648 & 0.691 & 0.543 & 0.315 & 0.509 & 0.271 & 0.694 & 0.777 & 0.555 & 0.719 & 0.697 & 0.586 \\
 4 & $\mathit{E_H}$ & 0.771 & 0.849 & 0.742 &  &  &  &  &  &  & 0.715 & 0.707 & 0.461 \\
 & $\mathit{E_{TH}}$ & 0.576 & 0.788 & 0.534 & 0.551 & 0.783 & 0.516 & 0.527 & 0.666 & 0.472 & 0.631 & 0.674 & 0.507 \\
 \hline
 \end{tabular}
 }
\caption{Accuracy scores of all models. \emph{Exp}: experiment number. \emph{Whole sample}: accuracy scores on the whole sample. \emph{Subset 1}: subset of transformed instances that have different gold label with respect to the control instances they were generated from. \emph{Subset 2}: subset of transformed instances that contain word pairs unseen at training time. \emph{Subset 3}: subset of control or transformed instances containing word pairs whose polarity does not match the instance's gold label.}
\label{table:models_accuracies} 
\end{table*}


\subsection{Transformation and Word Pairs}
\label{section:transformation_and_word_pairs}

Given a set of word pairs of the form $W=(w_1, w_2)$, where $w_1$ and $w_2$ hold under a semantic relation $s \in$ $\{antonymy,$ $hypernymy,$ $hyponymy\}$, we look through the training set for instances $i_k=(p_k, h_k)$, where $p_k$ and $h_k$ are premise and hypothesis sentences, respectively, such that $w_1 \in p_k \; \textrm{and} \; w_2 \in h_k$. For each instance $i_k$ we apply transformation $T$: we swap $w_1$ with $w_2$; this transformation yields an instance $i_m=(p_m, h_m)$ where $w_2 \in p_m, w_1 \in h_m \; \textrm{and} \; w_1 \notin p_m, w_2 \notin h_m$.\footnote{If a word $w_1$ or $w_2$ appears more than once, we replace all the appearances with its corresponding pair, $w_2$ or $w_1$.}

An example of transformation $T$ on a \emph{contradiction} instance $i_k$ is the following:

\begin{example}
$p_k:$ A soccer game occurring at sunset.  \\
$h_k:$ A basketball game is occurring at sunrise. \\
\label{example:instance_contradiction_inSitu}
\end{example}

Where the word pair $\mathit{(sunset, sunrise)}$ are antonyms. After applying transformation $T$, we obtain the new \emph{contradiction} instance $i_m$:

\begin{example}
$p_m:$ A soccer game occurring at sunrise. \\
$h_m:$ A basketball game is occurring at sunset. \\
\label{example:instance_contradiction_reversed_inSitu}
\end{example}

Consider now the following instance $i_l$ (class label \emph{entailment}): 

\begin{example}
$p_l:$ A little girl hugs her brother on a footbridge in a forest.  \\
$h_l:$ A pair of siblings are on a bridge. \\
\label{example:instance_entailment_inSitu}
\end{example}

If we now apply transformation $T$ on the hypernym word pair $\mathit{(footbridge,bridge)}$  we derive the new instance $i_n$ (class \emph{neutral}):

\begin{example}
$p_n:$ A little girl hugs her brother on a bridge in a forest. \\
$h_n:$ A pair of siblings are on a footbridge. \\
\label{example:instance_entailment_reversed_inSitu}
\end{example}

Since swapping word pairs under hypernymy or hyponymy relations may yield a different class label for the new instance, we manually annotate all the instances in the new sample, discarding those that are semantically incoherent.


\subsection{Experimental Conditions}
\label{section:experimental_conditions}

We consider two types of sentential context for the word pairs, namely \emph{in situ} and \emph{ex situ}. Examples of instances under the \emph{in situ} condition are Examples \ref{example:instance_contradiction_inSitu}, \ref{example:instance_contradiction_reversed_inSitu}, \ref{example:instance_entailment_inSitu}, and \ref{example:instance_entailment_reversed_inSitu} in Section \ref{section:transformation_and_word_pairs}. 
The name \emph{in situ} refers to the fact that we analyze the effect of the transformation $T$ within the original context of the premise and hypothesis sentences. This allows to control for confounding factors, such as sentence length and order of the context words.

We also consider an \emph{ex situ} condition in which we remove the word pair from the original premise and hypothesis and analyze the effect of the transformation $T$ within a simplified sentential context which is the same in premise and hypothesis. Specifically, we randomly select either the premise or hypothesis context from the original instance and copy it into both positions. In this way, we obtain a sentence pair where the only difference between the premise and hypothesis is the word pair, which allows us to isolate the effect of this pair from its interaction with the surrounding context; this condition thus allows to control for context words. This process yields a new set of instances, which we refer to as $E$.

An example of an \emph{ex situ} instance can be constructed from Example \ref{example:instance_contradiction_inSitu} (Section \ref{section:transformation_and_word_pairs}). If the premise sentence is selected, then after performing the procedure described above, the following sentence pair $e_k$ is generated:

\begin{example}
$p_k:$ A soccer game occurring at sunset. \\
$h_k:$ A soccer game occurring at sunrise. \\
\label{example:instance_contradiction_exSitu}
\end{example}

Given a sample $E$, we apply the transformation $T$ in order to generate a transformed sample $E_T$ where the word pairs are swapped, similar to the procedure applied in Section \ref{section:transformation_and_word_pairs} on SNLI control instances in order to generate their transformed instances counterpart. In the latter case, we say that given a sample of control instances $I$ we generate a transformed sample $I_T$.

As an example of obtaining a transformed \emph{ex situ} instance, we apply $T$ to $\mathit{(sunset, sunrise)}$ in Example \ref{example:instance_contradiction_exSitu} to obtain the new instance $e_m$:

\begin{example}
$p_m:$ A soccer game occurring at sunrise. \\
$h_m:$ A soccer game occurring at sunset. \\
\label{example:instance_contradiction_reversed_exSitu}
\end{example}

We note that for both conditions, \emph{in situ} and \emph{ex situ}, the same word pairs are swapped, so the differences are the surrounding context words and the factors being controlled.


\subsection{Test Sets}

In each experiment we use two sets of instances in order to measure the robustness of the models and analyze our target factors: 1) The control instances where the target word pair is in its original position and 2) the transformed instances generated after applying transformation $T$. The name of each set corresponds with the experimental setting it is used in. Samples used in \emph{in situ} experiments are named as $I$, and $E$ for \emph{ex situ}. Subscripts distinguish both the type of word pairs ($A$ for antonyms and $H$ for hypernym/hyponym) and the type of set (control or transformed). For example, $\mathit{I_A}$ refers to the control \emph{in situ} set whose instances contain antonym word pairs, whereas $\mathit{E_{TH}}$ refers to the \emph{ex situ} transformed test set containing hypernym/hyponym swapped word pairs.

We clarify: a) the sets $\mathit{I_A}$ and $\mathit{I_H}$ are sampled from the SNLI dataset; b) transformed test sets are generated from control sets containing control instances; c) we refer to the sets $\mathit{E_A}$ and $\mathit{E_H}$ as control test sets because the target word pairs are in their original position, and we apply $T$ on them in order to obtain the transformed samples $\mathit{E_{TA}}$ and $\mathit{E_{TH}}$, respectively.

Details about the sets: In order to build set $\mathit{I_A}$, we sample only \emph{contradiction} instances (instances in $\mathit{E_A}$ are also \emph{contradictions}). We use the antonym word pairs from \cite{mohammad2013computing} to yield the sets $\mathit{I_{TA1}}$ and $\mathit{E_{TA}}$, which also only contain \emph{contradictions} since the relation of antonymy is symmetric.\footnote{The word pair $\mathit{(sunset, sunrise)}$ holds in an antonymy relation regardless of the position of the words in premise and hypothesis sentences.} We build two more sets, $\mathit{I_{TA2}}$ and $\mathit{I_{TA3}}$ (explained in Section \ref{section:experiment1}). Sets $\mathit{I_H}$, $\mathit{E_H}$, $\mathit{I_{TH}}$, and $\mathit{E_{TH}}$ contain instances with any class label. In order to generate sets $\mathit{I_{TH}}$ and $\mathit{E_{TH}}$, we use the hypernym word pairs from \cite{baroni-EtAl:2012:EACL2012}. We manually annotate these transformed sets and discard incoherent instances.


\subsection{Factors Under Study}

We describe the three target factors that we hypothesize that affect the models' response. 

\paragraph{Insensitivity} is the name we give to the tendency of a model to predict the original label on a transformed instance that is similar to a control instance. Thus a model would be insensitive if, for example, it incorrectly predicts the same class label for both the control instance in Example \ref{example:instance_entailment_inSitu} and the transformed instance in Example \ref{example:instance_entailment_reversed_inSitu} just because they closely resemble each other. A simple measure of the impact of this effect is to look at the accuracy on the subset of instances in which the gold label was changed by the transformation. We show this effect by statistically correlating the rate of correct predictions with changes in the labels predicted. 

\paragraph{Unseen Word Pairs} are another factor we can use to evaluate robustness. In this case, we are interested in the subset of transformed instances where the swapped word pair is now in an order within premise and hypothesis that was unseen in the training data. An example is Example \ref{example:instance_contradiction_reversed_inSitu} which contains the unseen word pair $\mathit{(sunrise, sunset)}$; i.e., no instance in the training set contains the word \emph{sunrise} in the premise and the word \emph{sunset} in the hypothesis. Poor performance on this subset reflects an inability to exploit the symmetry (antonym pairs) or anti-symmetry (hypernym pairs) of the word pairs involved. We show models' abilities to cope with unseen pairs by statistically associating proportions of instances containing unseen pairs with incorrect predictions rates.

\paragraph{Polarity} is the name we give to the association between a word pair and the most frequent class it is found in across training instances. For example, we associate the word pair $\mathit{(sunset, sunrise)}$ with polarity \emph{contradiction} because it mainly appears on training instances with label \emph{contradiction}. We define four main categories of polarity: \emph{neutral}, \emph{contradiction}, \emph{entailment}, and \emph{none} for unseen word pairs.\footnote{We also define categories when a word pair appears the same number of times in two classes, such as \emph{entailment-neutral}, though these cases are rare.} 
Accuracy on the subset of instances where polarity and gold label disagree is an indicator of the extent to which a model is influenced by this factor. For example, a model incorrectly predicting label \emph{entailment} for the instance in Example \ref{example:instance_entailment_reversed_inSitu} (class \emph{neutral}) based on the polarity of class \emph{entailment} of its word pair $\mathit{(bridge, footbridge)}$ indicates that the model is influenced by this factor. We show this influence by statistically correlating labels predicted with polarities.


\section{Experiments and Results}

Table \ref{table:models_accuracies} presents the performance of the models across the different test sets. In general, DAM and ESIM seem to be more robust than CE, with the latter's accuracy degrading to essentially random performance on the most challenging subsets. However, this general trend is reversed in a single row of the table. On $\mathit{E_{TH}}$, ESIM shows a comparable performance to CE. And on Subset 3 of $\mathit{I_H}$, DAM appears to rely on a bias (polarity) in the same way as CE. Overall, all models are affected by the three target factors, dropping performance up to 0.25, 0.20, and 0.28 for ESIM, DAM, CE, respectively, just by virtue of our simple transformation of swapping words.


\subsection{Experiment 1: Swapping Antonyms in \emph{In Situ} Instances}
\label{section:experiment1}

In this experiment we use sets $\mathit{I_A}$ and $\mathit{I_{TA1}}$. Swapping antonyms seems to have no effect on the overall performance of the DAM model on $\mathit{I_{TA1}}$ when compared to $\mathit{I_A}$, and little effect on ESIM. Thus these two models appear to be robust to this transformation. Nonetheless, further analysis will not support the conclusion that both models have learned that antonymy is symmetric, and we will show that this seemingly robust behavior is due to confounding factors and not due to inference abilities. Accuracy scores of CE model seem to reveal that it is much less robust to the antonym swap, with performance significantly dropping by roughly 10.5\% according to a McNemar's test.

\paragraph{Insensitivity}

Because instances in $\mathit{I_{TA1}}$ are \emph{contradiction}, we perform a proxy experiment to understand the models' sensitivity. From $\mathit{I_A}$, we substitute one of the antonyms in each word pair (in each instance) with a hyponym, hypernym, or synonym\footnote{We manually select these from WordNet such that it appears at least $t=10$ times in the training set on either the premise sentences or the hypothesis sentences.} of the other. Doing this on both the premise and hypothesis yields two new samples, $\mathit{I_{TA2}}$ and $\mathit{I_{TA3}}$, which we manually annotate.

Examples of control (Example \ref{example:instance_contradiction_inSitu_2}) and transformed (Example \ref{example:instance_contradiction_fromAntonymToHypernym_inSitu}) instances are given below, showing the replacement of \emph{young}, in the hypothesis, with \emph{aged}, a synonym of \emph{elderly} from the premise. This transformation changes gold-label from \emph{contradiction} to \emph{neutral}. Approximately, half the sample yields such changes in gold-label.

\begin{example}
$p_k:$ An elderly woman sitting on a bench.  \\
$h_k:$ A young mother sits down. \\
\label{example:instance_contradiction_inSitu_2}
\end{example}

\begin{example}
$p_m:$ An elderly woman sitting on a bench.  \\
$h_m:$ An aged mother sits down. \\
\label{example:instance_contradiction_fromAntonymToHypernym_inSitu}
\end{example}

This transformation leads to a considerable drop in overall performance for all models when accuracy scores on sets $\mathit{I_{TA2}}$ and $\mathit{I_{TA3}}$ are compared to the accuracy on the control instances in $\mathit{I_A}$: up to 0.175 (CE), 0.201 (DAM), and 0.24 (ESIM) points (Table \ref{table:models_accuracies}). To test if insensitivity to the transformation is associated with these behaviors, we measure accuracy only on those instances that changed gold-label (Subset 1 from the sets $\mathit{I_{TA2}}$ and $\mathit{I_{TA3}}$), where we see a further reduction in performance for all models. 2-way tests of independence provide strong evidence for the insensitivity of the models (CE: $\chi^2(1)=73.33$, DAM: $\chi^2(1)=108.30$, ESIM: $\chi^2(1)=175.34$). 

Table \ref{table:ESIM_experiment1_insensitivity} shows the case for ESIM: most of its incorrect predictions are due to predicting the same label on both control and transformed instances when these two type of instances have different gold labels. Paradoxically, this effect works in the models' favour in the antonym swapping case ($\mathit{I_{TA1}}$) because all the gold-labels remain as \emph{contradiction}. Thus ignoring the transformation will avoid any loss in performance.

\begin{table}[ht!]
\centering
\resizebox{\columnwidth}{!}{
\scalebox{1.0}{
\begin{tabular}{c c c }
 \hline
  & \multicolumn{2}{c}{Distribution of predictions} \\[0.5ex] 
 Labels predicted & correct & incorrect \\ 
 \hline\hline
 change & 155 & 31 \\ 
 no change & 8 & 100 \\
 \hline
 \end{tabular}
}
}
\caption{Contingency table for ESIM: Predictions on transformed instances with different gold labels from those of the control instances.}
\label{table:ESIM_experiment1_insensitivity}
\end{table}

\paragraph{Unseen Word Pairs}

The results in the column Subset 2 of $\mathit{I_{TA1}}$ (Table \ref{table:models_accuracies}) suggest that performance on unseen word pairs is weak. However, only 40 instances within $\mathit{I_{TA1}}$ contain unseen antonym pairs; thus the impact of this result may be limited. 2-way tests of homogeneity show that the difference in accuracy of predictions in instances containing seen or unseen word pairs is nonetheless significant for all models (CE: $\chi^2(1)=19.46$, DAM: $\chi^2(1)=74.16$, ESIM: $\chi^2(1)=39.33$). In other words, the models struggle to recognize the reversed antonym pairs, even though they were all seen in their original order at training time. This effect can be seen, for example, in the contingency table for DAM in Table \ref{table:DAM_experiment1_unseenWordPairs}.

\begin{table}[ht!]
\centering
\resizebox{\columnwidth}{!}{
\scalebox{0.07}{
\begin{tabular}{c c c }
 \hline
  & \multicolumn{2}{c}{Word pairs} \\[0.5ex] 
 Predictions & seen & unseen \\ 
 \hline\hline
 correct & 567 & 20 \\ 
 incorrect & 13 & 20 \\
 \hline
 \end{tabular}
}
}
\caption{Contingency table for DAM: Predictions distributed according to instances containing a seen or an unseen antonym word pair.}
\label{table:DAM_experiment1_unseenWordPairs}
\end{table}

\paragraph{Polarity}

Only 11\% of the instances in the transformed sample $\mathit{I_{TA1}}$ contain word pairs that have polarity other than \emph{contradiction}. Thus, a model relying only on this factor could achieve an accuracy of 89\%. We investigate if the predicted labels on instances in $\mathit{I_{TA1}}$ are associated with the polarity of the transformed word pair. For all models, independence tests are highly significant (CE: $\chi^2(6)=30.69$, DAM: $\chi^2(6)=101.26$, ESIM: $\chi^2(6)=64.40$). Table \ref{table:DAM_experiment1_wordPairPolarity} shows that the predictions of DAM change according to the polarity of the word pairs. For example, when the polarity is \emph{contradiction}, around 98.5\% of the predictions are \emph{contradictions}; however, this figure changes when the polarity is \emph{neutral} where the rate of correct predictions (\emph{contradictions}) fall to 80.7\%, and a more dramatic fall is observed when the word pairs are unseen (polarity \emph{none}) where only 50\% of the predictions are correct. This is strong evidence that the models learned to rely on polarity.

We note that a model with perfect accuracy on $\mathit{I_{TA1}}$, would lead to a statistic that does not reject the null hypothesis, showing in this case that the predictions are independent of polarity.

\begin{table}[ht!]
\centering
\resizebox{\columnwidth}{!}{
\scalebox{1.0}{
\begin{tabular}{c c c c}
 \hline
 \diagbox{Polarity}{Prediction} & Neutral & Contradiction & Entailment \\ [0.5ex]
 \hline\hline
 Neutral & 5 & 21 & 0 \\ 
 Contradiction & 5 & 543 & 3 \\
 Entailment & 0 & 3 & 0 \\
 None & 8 & 20 & 12 \\
 \hline
 \end{tabular}
}
}
\caption{Contingency table for DAM: Predictions distributed according to the polarity of target word pairs found in the transformed instances.}
\label{table:DAM_experiment1_wordPairPolarity}
\end{table}


\subsection{Experiment 2: Swapping Antonyms in \emph{Ex Situ} Instances}

In this experiment, we use samples $\mathit{E_A}$ and $\mathit{E_{TA}}$. Swapping antonyms has little effect on the performance of all models, where the biggest drop comes from DAM (0.029 points). However, the CE model performs quite poorly at both samples (0.508 and 0.48 accuracy points on $\mathit{E_A}$ and $\mathit{E_{TA}}$); this drop in performance, with respect to the \emph{in situ} condition, suggests that the repeated sentence context is too different from the structure of the training instances for the CE model to generalize effectively. 

In this condition, we refrain from analyzing the effect of insensitivity, since doing so would require a transformation similar to that in the \emph{in situ} condition, which might add an extra layer of change and the results may turn difficult to interpret. 

\paragraph{Unseen Word Pairs}

Accuracy scores strongly suggest that the models are weak at dealing with unseen antonym pairs (Subset 2 of $\mathit{E_{TA}}$ in Table \ref{table:models_accuracies}); drops in performance on this subset range from 0.315 up to 0.429 points across the three models. Tests of homogeneity show strong evidence of this weakness for all models (CE: $\chi^2(1)=15.91$, DAM: $\chi^2(1)=59.17$, ESIM: $\chi^2(1)=44.72$). Comparing results on this subset with those of Subset 2 in $\mathit{I_{TA1}}$, we notice that ESIM and DAM keep similar behavior, but CE seems to be strongly affected by this context type.

\paragraph{Polarity}

All models perform poorly in the subset of instances where polarity disagrees with gold label of the instance (Subset 3 of $\mathit{E_{TA}}$), showing that the models' behavior rely on this bias. 
These results are highly significant (CE: $\chi^2(6)=34.37$, DAM: $\chi^2(6)=136.99$, ESIM: $\chi^2(6)=103.47$). This is further evidence that the models get \emph{confused} with a simple reversal of an antonym pair. 


\subsection{Experiment 3: Swapping Hypernyms and Hyponyms in \emph{In Situ} Instances}

We now study the effect on the robustness of the systems when we swap hypernym and hyponym word pairs in \emph{in situ} instances. Whole sample accuracy scores in Table \ref{table:models_accuracies} significantly drop, according to McNemar's tests, by 0.25 (ESIM), 0.285 (CE), and 0.128 (DAM) points when we compare scores on control instances ($\mathit{I_H}$) with those on transformed instances ($\mathit{I_{TH}}$). We investigate the role of our target factors on these behaviors.

\paragraph{Insensitivity}

Around 42\% of the instances in $\mathit{I_{TH}}$ (Subset 1) have different gold label from those in $\mathit{I_H}$. On these instances, the models' results are severely impaired: CE and ESIM models' performances drop to close-to-random (0.271 and 0.315), while DAM decreases by 0.18 points. All models' errors on this subset are strongly associated with failure to change the predicted class (CE:$\chi^2(1)=90.73$, DAM:$\chi^2(1)=101.52$, ESIM:$\chi^2(1)=150.92$). In contrast to the case in Experiment 1, insensitivity acts in detriment of the models' robustness when gold labels change after the transformation.

\paragraph{Unseen Word Pairs}

Whereas model performance was significantly worse on unseen antonym pairs, this effect is not obvious on the hyponym-hypernym results (Subset 2 of $\mathit{I_{TH}}$). In fact, all models have a slightly higher accuracy on this subset than overall. Homogeneity tests find no evidence of an association between unseen word pairs and incorrect predictions for any model (CE:$\chi^2(1)=0.00036, p=0.98$, DAM:$\chi^2(1)=0.98, p=0.32$, ESIM:$\chi^2(1)=0.178, p=0.67$). This effect may be explained by the models exploiting information from word embeddings. It has been shown that word embeddings are able to capture hypernymy \cite{sanchez-riedel:2017:EACLshort}; thus the models may use this information to generalize to unseen hypernym pairs.

\paragraph{Polarity}

We find very strong evidence for an association between polarity and class label predicted on sample $\mathit{I_H}$ for all models (CE:$\chi^2(10)=168.40$, DAM:$\chi^2(10)=182.76$, ESIM:$\chi^2(10)=157.76$). However, for sample $\mathit{I_{TH}}$, only DAM keeps this strong correlation ($\chi^2(14)=47.71$). In the case of CE, we find weak evidence in favour of this correlation on instances of $\mathit{I_{TH}}$ ($\chi^2(14)=25.27, p=0.03$). For ESIM we find no evidence of correlation ($\chi^2(14)=22.72, p=0.06$), thus we do not reject the null hypothesis. Polarity's influence can be observed in Subset 3 of $\mathit{I_H}$ (Table \ref{table:models_accuracies}), where we observe a drop in accuracy for instances whose gold labels do not match the polarity of the word pairs, compared to the accuracy of the whole sample; this means that when the models have polarity as a cue, they improve performance. 


\subsection{Experiment 4: Swapping Hypernyms and Hyponyms in \emph{Ex Situ} Instances}

All models' performance significantly drop ($p<0.01$) after our transformation by 0.208 (CE), 0.061 (DAM) and 0.195 (ESIM) points, where performance of ESIM is comparable to that of CE on both samples, $\mathit{E_H}$ and $\mathit{E_{TH}}$. Compared to the \emph{in situ} condition, DAM's performance improves, opposite to CE's and ESIM's behavior.

\paragraph{Insensitivity}

The drop in performance described above can be partially explained by insensitivity to changes in gold label, since around 93\% of the instances in $\mathit{E_{TH}}$ changed gold-label with respect to $\mathit{E_H}$. We find strong statistical evidence for this hypothesis (CE:$\chi^2(1)=175.19$, DAM:$\chi^2(1)=158.62$, ESIM:$\chi^2(1)=252.27$). However, in the case of DAM, this factor seems to play a small role on its behavior as seen when we compare accuracy on Subset 1 with that of the whole transformed sample. 

Insensitivity seems to have a bigger influence on the models when the transformed instances are closer to the training set: Accuracy scores on Subset 1 from $\mathit{I_{TH}}$ are smaller than those on Subset 1 from $\mathit{E_{TH}}$. 

\paragraph{Unseen Word Pairs}

Similar to the \emph{in situ} condition, our homogeneity tests show no evidence for incorrect predictions being due to unseen word pairs (CE:$\chi^2(1)=0.35, p=0.55$, DAM:$\chi^2(1)=2.43, p=0.11$, ESIM:$\chi^2(1)=0.183, p=0.66$). We posit the same explanation as before: Models may use hypernymy information contained in the embeddings.

\paragraph{Polarity}

We find statistically high correlation of the models' predictions with the polarity of the word pairs in the instances from both samples, $\mathit{E_H}$ (CE:$\chi^2(10)=261.77$, DAM:$\chi^2(10)=312.67$, ESIM:$\chi^2(10)=176.38$) and $\mathit{E_{TH}}$ (CE:$\chi^2(14)=56.52$, DAM:$\chi^2(14)=258.09$, ESIM:$\chi^2(10)=105.70$). This evidence indicates that all models use, to some extent, the polarity as a feature for predicting class labels.


\section{Discussion and Conclusions}

Although all three models achieve strong results on the original SNLI development set (CE: 0.782, DAM: 0.854, ESIM: 0.882), each model exhibits particular weaknesses on the transformed training instances. 
Notably, all perform poorly on $\mathit{I_{TH}}$ instances in which the gold label is changed, with ESIM and CE performing below the level of chance. 
Thus, on these instances, 
the models tend to predict the label of the original unaltered training instance
and inference in this case is similar to nearest-neighbour prediction. 

On the other hand, much better performance is obtained for the DAM and ESIM models on $\mathit{I_{TH}}$ instances containing unseen word pairs, 
indicating these models have learned to infer hypernym/hyponym relations from information in the pre-trained word embeddings. 
In contrast, performance on the unseen word pairs in $\mathit{I_{TA1}}$ and $\mathit{E_{TA}}$ suggests that inferring antonymy from the embeddings is more difficult.

Weak performance is seen again on the $E_{A}$ and $\mathit{E_{TA}}$ instances where the polarity of the antonym pair is not consistent with the gold label. 
For these cases, the only difference between premise and hypothesis is the antonym pair, 
and the models tend to fall back on predicting the most frequent gold label seen for that word pair.

One result that remains anomalous is the overall performance of the ESIM model on the whole $\mathit{E_{TH}}$ sample.
While this sample contains unseen word pairs and instances in which the gold label changes or is inconsistent with polarity, these effects do not by themselves explain the poor performance overall.
Neither is this weakness explained by the \emph{ex situ} structure, in which premise and hypothesis differ by only one word, as performance on the control \emph{ex situ} sample, $E_{H}$, is much stronger. 
The effect, then, appears to be due to an interaction of the \emph{ex situ} structure in combination with the transformation.

In the present work, we have limited ourselves to examining single influences independently. 
However, there are undoubtedly manifold interactions contributing to model performance. 
In fact, the complexities of these models (LSTMs, attention mechanisms and MLPs) are specifically intended 
to capture the interactions between the words in the premise and hypothesis. 
Further work is required to understand what these interactions are and how they contribute to performance.
Fully uncovering these factors in current NLI datasets is a pre-requisite for the construction of 
more effective resources in the future.

\section*{Acknowledgments}
We thank Raul Ortiz Pulido and Erick Sanchez Carmona for insightful discussions, Pasquale Minervini for providing the implementations of DAM and ESIM, Pontus Stenetorp for providing valuable feedback on the manuscript, and Johannes Welbl for insightful comments. The first author was recipient of a scholarship from CONACYT. This work was supported by an Allen Distinguished Investigator Award and the EU H2020 SUMMA project (grant agreement number 688139).


\bibliography{my_naacl_ref}

\begin{thebibliography}{}
\expandafter\ifx\csname natexlab\endcsname\relax\def\natexlab#1{#1}\fi

\bibitem[{Alpaydin(2010)}]{Alpaydin:2010:IML:1734076}
Ethem Alpaydin. 2010.
\newblock {\em Introduction to Machine Learning\/}.
\newblock The MIT Press, 2nd edition.

\bibitem[{B.~Hashemi and Hwa(2016)}]{bhashemi-hwa:2016:EMNLP2016}
Homa B.~Hashemi and Rebecca Hwa. 2016.
\newblock An evaluation of parser robustness for ungrammatical sentences.
\newblock In {\em Proceedings of the 2016 Conference on Empirical Methods in
  Natural Language Processing\/}. Association for Computational Linguistics,
  Austin, Texas, pages 1765--1774.

\bibitem[{Baroni et~al.(2012)Baroni, Bernardi, Do, and
  Shan}]{baroni-EtAl:2012:EACL2012}
Marco Baroni, Raffaella Bernardi, Ngoc-Quynh Do, and Chung-chieh Shan. 2012.
\newblock Entailment above the word level in distributional semantics.
\newblock In {\em Proceedings of the 13th Conference of the European Chapter of
  the Association for Computational Linguistics\/}. Association for
  Computational Linguistics, Avignon, France, pages 23--32.

\bibitem[{Birkett and Newton-Fisher(2011)}]{birkett2011abnormal}
Lucy~P. Birkett and Nicholas~E. Newton-Fisher. 2011.
\newblock How abnormal is the behaviour of captive, zoo-living chimpanzees?
\newblock {\em PLOS ONE\/} 6(6):1--7.

\bibitem[{Bolukbasi et~al.(2016)Bolukbasi, Chang, Zou, Saligrama, and
  Kalai}]{bolukbasi-nips16}
Tolga Bolukbasi, Kai-Wei Chang, James~Y. Zou, Venkatesh Saligrama, and Adam~T.
  Kalai. 2016.
\newblock Man is to computer programmer as woman is to homemaker? debiasing
  word embeddings.
\newblock In D.~D. Lee, M.~Sugiyama, U.~V. Luxburg, I.~Guyon, and R.~Garnett,
  editors, {\em Advances in Neural Information Processing Systems 29\/}, Curran
  Associates, Inc., pages 4349--4357.

\bibitem[{Bowman et~al.(2015)Bowman, Angeli, Potts, and
  Manning}]{snli:emnlp2015}
Samuel~R. Bowman, Gabor Angeli, Christopher Potts, and Christopher~D. Manning.
  2015.
\newblock A large annotated corpus for learning natural language inference.
\newblock In {\em Proceedings of the 2015 Conference on Empirical Methods in
  Natural Language Processing\/}. Association for Computational Linguistics,
  Lisbon, Portugal, pages 632--642.

\bibitem[{Chen et~al.(2017)Chen, Zhu, Ling, Wei, Jiang, and
  Inkpen}]{chen-EtAl:2017:Long3}
Qian Chen, Xiaodan Zhu, Zhen-Hua Ling, Si~Wei, Hui Jiang, and Diana Inkpen.
  2017.
\newblock Enhanced lstm for natural language inference.
\newblock In {\em Proceedings of the 55th Annual Meeting of the Association for
  Computational Linguistics (Volume 1: Long Papers)\/}. Association for
  Computational Linguistics, Vancouver, Canada, pages 1657--1668.

\bibitem[{Craven and Shavlik(1996)}]{craven1996extracting}
Mark Craven and Jude~W. Shavlik. 1996.
\newblock Extracting tree-structured representations of trained networks.
\newblock In D.~S. Touretzky, M.~C. Mozer, and M.~E. Hasselmo, editors, {\em
  Advances in Neural Information Processing Systems 8\/}, MIT Press, pages
  24--30.

\bibitem[{Crawford and Calo(2016)}]{crawford2016there}
Kate Crawford and Ryan Calo. 2016.
\newblock There is a blind spot in ai research.
\newblock {\em Nature\/} 538(7625).

\bibitem[{Dagan et~al.(2009)Dagan, Dolan, Magnini, and
  Roth}]{dagan_dolan_magnini_roth_2009}
Ido Dagan, Bill Dolan, Bernardo Magnini, and Dan Roth. 2009.
\newblock Recognizing textual entailment: Rational, evaluation and approaches.
\newblock {\em Natural Language Engineering\/} 15(4):i--xvii.

\bibitem[{Dagan and Glickman(2004)}]{dagan_2004}
Ido Dagan and Oren Glickman. 2004.
\newblock Probabilistic textual entailment: generic applied modeling of
  language variability.
\newblock In {\em PASCAL Workshop on Learning Methods for Text Understanding
  and Mining\/}. Grenoble, France.

\bibitem[{Epling and Pierce(1986)}]{Epling1986}
W.~Frank Epling and W.~David Pierce. 1986.
\newblock The basic importance of applied behavior analysis.
\newblock {\em The Behavior Analyst\/} 9(1):89--99.

\bibitem[{Isabelle et~al.(2017)Isabelle, Cherry, and
  Foster}]{isabelle-cherry-foster:2017:EMNLP2017}
Pierre Isabelle, Colin Cherry, and George Foster. 2017.
\newblock A challenge set approach to evaluating machine translation.
\newblock In {\em Proceedings of the 2017 Conference on Empirical Methods in
  Natural Language Processing\/}. Association for Computational Linguistics,
  Copenhagen, Denmark, pages 2476--2486.

\bibitem[{Jia and Liang(2017)}]{jia-liang:2017:EMNLP2017}
Robin Jia and Percy Liang. 2017.
\newblock Adversarial examples for evaluating reading comprehension systems.
\newblock In {\em Proceedings of the 2017 Conference on Empirical Methods in
  Natural Language Processing\/}. Association for Computational Linguistics,
  Copenhagen, Denmark, pages 2011--2021.

\bibitem[{Kummerfeld et~al.(2012)Kummerfeld, Hall, Curran, and
  Klein}]{kummerfeld-EtAl:2012:EMNLP-CoNLL}
Jonathan~K. Kummerfeld, David Hall, James~R. Curran, and Dan Klein. 2012.
\newblock Parser showdown at the wall street corral: An empirical investigation
  of error types in parser output.
\newblock In {\em Proceedings of the 2012 Joint Conference on Empirical Methods
  in Natural Language Processing and Computational Natural Language
  Learning\/}. Association for Computational Linguistics, Jeju Island, Korea,
  pages 1048--1059.

\bibitem[{Li et~al.(2016)Li, Monroe, and Jurafsky}]{DBLP:journals/corr/LiMJ16a}
Jiwei Li, Will Monroe, and Dan Jurafsky. 2016.
\newblock Understanding neural networks through representation erasure.
\newblock {\em CoRR\/} abs/1612.08220.

\bibitem[{LoBue and Yates(2011)}]{LoBue:2011:TCK:2002736.2002805}
Peter LoBue and Alexander Yates. 2011.
\newblock Types of common-sense knowledge needed for recognizing textual
  entailment.
\newblock In {\em Proceedings of the 49th Annual Meeting of the Association for
  Computational Linguistics: Human Language Technologies: Short Papers - Volume
  2\/}. Association for Computational Linguistics, Stroudsburg, PA, USA, HLT
  '11, pages 329--334.

\bibitem[{Maccartney(2009)}]{NLI}
Bill Maccartney. 2009.
\newblock {\em Natural Language Inference\/}.
\newblock Ph.D. thesis, Stanford, CA, USA.
\newblock AAI3364139.

\bibitem[{McDonald(2014)}]{mcdonald-2014}
J.H. McDonald. 2014.
\newblock {\em Handbook of Biological Statistics (3rd ed.)\/}.
\newblock Sparky House Publishing, Baltimore, Maryland.

\bibitem[{Mench(1998)}]{doi:10.1093/ilar.39.1.20}
Joy Mench. 1998.
\newblock Why it is important to understand animal behavior.
\newblock {\em ILAR Journal\/} 39(1):20--26.

\bibitem[{Mohammad et~al.(2013)Mohammad, Dorr, Hirst, and
  Turney}]{mohammad2013computing}
Saif~M. Mohammad, Bonnie~J. Dorr, Graeme Hirst, and Peter~D. Turney. 2013.
\newblock Computing lexical contrast.
\newblock {\em Computational Linguistics\/} 39(3):555--590.

\bibitem[{Oliphant(2007)}]{4160250}
T.~E. Oliphant. 2007.
\newblock Python for scientific computing.
\newblock {\em Computing in Science Engineering\/} 9(3):10--20.

\bibitem[{Parikh et~al.(2016)Parikh, T\"{a}ckstr\"{o}m, Das, and
  Uszkoreit}]{parikh-EtAl:2016:EMNLP2016}
Ankur Parikh, Oscar T\"{a}ckstr\"{o}m, Dipanjan Das, and Jakob Uszkoreit. 2016.
\newblock A decomposable attention model for natural language inference.
\newblock In {\em Proceedings of the 2016 Conference on Empirical Methods in
  Natural Language Processing\/}. Association for Computational Linguistics,
  Austin, Texas, pages 2249--2255.

\bibitem[{Ribeiro et~al.(2016)Ribeiro, Singh, and
  Guestrin}]{Ribeiro:2016:WIT:2939672.2939778}
Marco~Tulio Ribeiro, Sameer Singh, and Carlos Guestrin. 2016.
\newblock "why should i trust you?": Explaining the predictions of any
  classifier.
\newblock In {\em Proceedings of the 22Nd ACM SIGKDD International Conference
  on Knowledge Discovery and Data Mining\/}. ACM, New York, NY, USA, KDD '16,
  pages 1135--1144.

\bibitem[{Sammons et~al.(2010)Sammons, Vydiswaran, and
  Roth}]{Sammons:2010:ATE:1858681.1858803}
Mark Sammons, V.~G.~Vinod Vydiswaran, and Dan Roth. 2010.
\newblock "ask not what textual entailment can do for you...".
\newblock In {\em Proceedings of the 48th Annual Meeting of the Association for
  Computational Linguistics\/}. Association for Computational Linguistics,
  Stroudsburg, PA, USA, ACL '10, pages 1199--1208.

\bibitem[{Sanchez and Riedel(2017)}]{sanchez-riedel:2017:EACLshort}
Ivan Sanchez and Sebastian Riedel. 2017.
\newblock How well can we predict hypernyms from word embeddings? a
  dataset-centric analysis.
\newblock In {\em Proceedings of the 15th Conference of the European Chapter of
  the Association for Computational Linguistics: Volume 2, Short Papers\/}.
  Association for Computational Linguistics, Valencia, Spain, pages 401--407.

\bibitem[{Seabold and Perktold(2010)}]{seabold2010statsmodels}
Skipper Seabold and Josef Perktold. 2010.
\newblock Statsmodels: Econometric and statistical modeling with python.
\newblock In {\em 9th Python in Science Conference\/}.

\bibitem[{Soman(2001)}]{soman2001effects}
Dilip Soman. 2001.
\newblock Effects of payment mechanism on spending behavior: The role of
  rehearsal and immediacy of payments.
\newblock {\em Journal of Consumer Research\/} 27(4):460--474.

\bibitem[{White et~al.(2017)White, Rastogi, Duh, and
  Van~Durme}]{white-EtAl:2017:I17-1}
Aaron~Steven White, Pushpendre Rastogi, Kevin Duh, and Benjamin Van~Durme.
  2017.
\newblock Inference is everything: Recasting semantic resources into a unified
  evaluation framework.
\newblock In {\em Proceedings of the Eighth International Joint Conference on
  Natural Language Processing (Volume 1: Long Papers)\/}. Asian Federation of
  Natural Language Processing, Taipei, Taiwan, pages 996--1005.

\bibitem[{Zhao et~al.(2017)Zhao, Wang, Yatskar, Ordonez, and
  Chang}]{zhao-EtAl:2017:EMNLP20173}
Jieyu Zhao, Tianlu Wang, Mark Yatskar, Vicente Ordonez, and Kai-Wei Chang.
  2017.
\newblock Men also like shopping: Reducing gender bias amplification using
  corpus-level constraints.
\newblock In {\em Proceedings of the 2017 Conference on Empirical Methods in
  Natural Language Processing\/}. Association for Computational Linguistics,
  Copenhagen, Denmark, pages 2979--2989.

\end{thebibliography}
\bibliographystyle{acl_natbib}

\end{document}